\documentclass[acmsmall]{acmart}

\AtBeginDocument{%
  \providecommand\BibTeX{{%
    \normalfont B\kern-0.5em{\scshape i\kern-0.25em b}\kern-0.8em\TeX}}}

\copyrightyear{2023} 
\acmYear{2023} 
\setcopyright{rightsretained} 
\acmConference[IUI '23 Companion]{28th International Conference on Intelligent User Interfaces}{March 27--31, 2023}{Sydney, NSW, Australia}
\acmBooktitle{28th International Conference on Intelligent User Interfaces (IUI '23 Companion), March 27--31, 2023, Sydney, NSW, Australia}
\acmDOI{10.1145/3581754.3584136}
\acmISBN{979-8-4007-0107-8/23/03}

\usepackage{caption}
\usepackage{subcaption}

\usepackage{xcolor}
\definecolor{darkgreen}{HTML}{005e19}
\definecolor{darkblue}{HTML}{240394}
\newcommand{\example}[1]{\textcolor{darkblue}{\textbf{\small{#1}}}}

\begin{document}

\title[Supporting Qualitative Analysis with Large Language Models]{Supporting Qualitative Analysis with Large Language Models: Combining Codebook with GPT-3 for Deductive Coding}

\author{Ziang Xiao}
\email{ziangxiao@microsoft.com}
\affiliation{%
  \institution{Microsoft Research}
  \city{Montreal}
  \state{Quebec}
  \country{Canada}
}
\author{Xingdi Yuan}
\email{eric.yuan@microsoft.com}
\affiliation{%
  \institution{Microsoft Research}
  \city{Montreal}
  \state{Quebec}
  \country{Canada}
}
\author{Q. Vera Liao}
\email{veraliao@microsoft.com}
\affiliation{%
  \institution{Microsoft Research}
  \city{Montreal}
  \state{Quebec}
  \country{Canada}
}

\author{Rania Abdelghani}
\email{rania.abdelghani@inria.fr}
\affiliation{%
  \institution{Inria} 
  \country{France}
}
\author{Pierre-Yves Oudeyer}
\email{pierre-yves.oudeyer@inria.fr}
\affiliation{%
  \institution{Inria} 
  \country{France}
}

\renewcommand{\shortauthors}{Xiao et al.}

\begin{abstract}
Qualitative analysis of textual contents unpacks rich and valuable information by assigning labels to the data. However, this process is often labor-intensive, particularly when working with large datasets. While recent AI-based tools demonstrate utility, researchers may not have readily available AI resources and expertise, let alone be challenged by the limited generalizability of those task-specific models. In this study, we explored the use of large language models (LLMs) in supporting deductive coding, a major category of qualitative analysis where researchers use pre-determined codebooks to label the data into a fixed set of codes. Instead of training task-specific models, a pre-trained LLM could be used directly for various tasks without fine-tuning through prompt learning. Using a curiosity-driven questions coding task as a case study, we found, by combining GPT-3 with expert-drafted codebooks, our proposed approach achieved fair to substantial agreements with expert-coded results. We lay out challenges and opportunities in using LLMs to support qualitative coding and beyond.
\end{abstract}



\maketitle

\section{Introduction}
Qualitative coding is a method of qualitative analysis that is used to identify patterns and categories in qualitative data, such as social media posts, open-ended survey responses, and field notes. Often, scientists will use coded data to derive theory or build models to understand the observed phenomenon further~\cite{hsieh2005three}. Qualitative coding can be a challenging process because it requires the researcher to understand and analyze data that is often complex, nuanced, and open to multiple interpretations. Multiple researchers need to spend a significant amount of time and effort to review the data, develop a codebook that can accurately describe each label, and iteratively code the data until reaching a reasonable inter-rater agreement. It is particularly hard when working with large datasets. In this study, we explore a novel approach that leverages large language models (LLMs) to facilitate qualitative coding.

Researchers have built AI-based tools to assist qualitative analysis~\cite{10.1145/3411764.3445591, liew2014optimizing, muller2016machine,paredes2017inquire}. These tools use natural language processing (NLP) and machine learning (ML) algorithms to help researchers identify patterns and themes in qualitative data. Two categories of models are often used, 1) unsupervised models, e.g., topic models, to help researchers discover themes, or 2) supervised models, e.g., logistic regression, to classify data into labels. However, both methods face their own challenges. Unsupervised models are difficult to steer. It is difficult for researchers to use those models for their complex or nuanced research questions. And supervised models often require high-quality large datasets or computing resources to achieve reasonable performance. Therefore, most of today's qualitative coding still relies on manual effort. 

The recent advent of LLMs (e.g., GPT-3~\cite{brown2020gpt3}, PaLM~\cite{chowdhery2022palm}, OPT~\cite{zhang2022opt}) offers new capabilities, including creative writing \cite{yuan2022wordcraft}, programming \footnote{https://github.com/features/copilot}, etc. Unlike traditional task-specific models, LLMs accept natural language prompts as input to perform various tasks~\cite{liu2021pre}. For example, LLMs could perform classification tasks if the prompt specifies a set of labels as the output space~\cite{gao2021making}. Compared to unsupervised models, the prompt could include specific instructions and examples (e.g., a codebook) to increase LLMs' performance in a new task with unseen data in a zero-shot/few-shot fashion. Prior work shows that LLMs can be prompted to boost an NLG system's performance by selecting better outputs from sampled candidates, according to some pre-defined metrics~\cite{yuan2022selecting}. Since LLMs operate effectively with natural language input and do not require training datasets, they lower the barriers for researchers without extensive AI expertise or resources to leverage AI capabilities in data analysis (although many of today's LLMs are proprietary).

Although recent studies have demonstrated LLMs utility in many domains, studies have shown LLMs are error-prone and have limited capability in capturing language structure and nuanced meanings ~\cite{korngiebel2021considering}, which are crucial in qualitative analysis. Therefore, in this preliminary study, we asked two research questions,

\begin{itemize}
        \item {\textbf{RQ1:} To what extent does our LLM-based approach agree with experts in a deductive coding task?}
        
        \item{\textbf{RQ2:} How do different prompt designs affect the coding results?}
    
\end{itemize}

We examined LLMs' capability in facilitating a deductive coding task where we combined GPT-3 with expert-developed codebooks to analyze children's curiosity-driven questions in terms of question complexity and syntactic structure. We found our proposed approach achieved fair to substantial agreements with experts (Question complexity: Cohen's $\kappa = 0.61$; Syntactic Structure: Cohen's $\kappa = 0.38$). Based on our preliminary findings, we present challenges and opportunities for utilizing LLMs in qualitative analysis.

\section{Deductive Coding Task}
We selected a deductive coding task as our starting point. The goal of deductive coding is to label the data based on a codebook. It is a top-down process where the researchers start by developing a codebook with an initial set of codes along with descriptions and examples based on the research focus or theory \cite{hsieh2005three}. Note that a coding process often involves both inductive coding (e.g. developing the code book) and deductive coding (coding all the data according to a code book). Our approach can be used to complete the second part or combined with the first part to help coders rapidly iterate on their codebook. 


\subsection{Case Study: Curiosity-driven questions Analysis}
We examined our approach in analyzing children's curiosity-driven questions. Understanding children's ability to ask curiosity-driven questions informs psychologists of a child's learning stage. We looked at two dimensions of a curiosity-driven question, question complexity and syntactic structure. The question complexity looks at if the answer to a question is a simple fact (e.g., ``How big is a dinosaur?'') or requires explaining a mechanism, a relationship, etc.(e.g., ``Why were dinosaurs so big?'')~\cite{wilen1991questioning}. The syntactic structure has four categories, 1) `closed' or declarative questions (e.g., ``Dinosaurs were big?''), 2) questions with questioning words in the middle of the sentence (e.g., ``The dinosaurs were how big?''), 3) questions without an interrogative formulation (e.g., ``Why the dinosaurs are big?''), and 4) questions with a questioning word at the beginning of the sentence that has interrogative syntax (e.g., ``Why are dinosaurs big?''). 

We collected a dataset with 668 children's questions in French \cite{abdelghani2022conversational} \footnote{For question complexity, we first used GPT-3 to translate questions into English. For the  syntactic structure, we kept questions in French to preserve its syntactic structure.}. A team of psychologists has developed a codebook and coded each question on the dimension of question complexity and syntactic structure. Our chosen dimensions cover two main categories of deductive coding, binary coding, and multi-level coding, with one focusing on the semantic meaning and the other looking at the syntactic structure. Additionally, the dataset and codebook have never been published online which is unseen by LLMs.

\section{GPT-3 Setup and Prompt Design}
We chose GPT-3 (davinci-text-002) with a temperature of 0.0 during the prompting process, because it was the most advanced version of GPT-3 that was publicly available when we conducted the experiments and 0.0 temperature (greedy decoding) guarantees the reproducibility of this study.

We explored two design dimensions of the prompt,
\begin{itemize}
    \item Codebook-centered vs. Example-centered: This dimension regards the structure of a prompt. In the codebook-centered prompt, we designed the prompt similar to how researchers read a codebook. The prompt follows the structure of [Code/ Description/ Examples]. For example, \example{Code: HIGH; Description: the answer to this question is not a simple fact but requires explaining a mechanism, a relationship, etc.; Examples: Why were dinosaurs so big?} The example-centered approach is inspired by the in-context learning in recent LLM works where the prompt explains the rationale behind each example \cite{liu2021pre}. For example, \example{``Why were dinosaurs so big?'' is an example of ``HIGH'' because the answer to this question is not a simple fact but requires explaining a mechanism, a relationship, etc.} The code, examples, and descriptions are the same for both designs.
    \item Zero-shot vs. One-shot vs. Few-shots: Since recent work showed conflicting results on the number of examples in a prompt \cite{liu2021pre}, we explored different prompt settings. In the Zero-shot setting, we give only Code and Description in Codebook-centered prompts \footnote{Since the example-centered approach requires at least one example, we did not have the zero-shot setting for the example-centered approach}. For the One-shot setting, we provided only one example for each code. And for the few-shot setting, we provided five examples for each code.   
\end{itemize}
\begin{figure*}[t]
         \begin{subfigure}[b]{1\textwidth}
            \centering
            \includegraphics[width=0.6\textwidth]{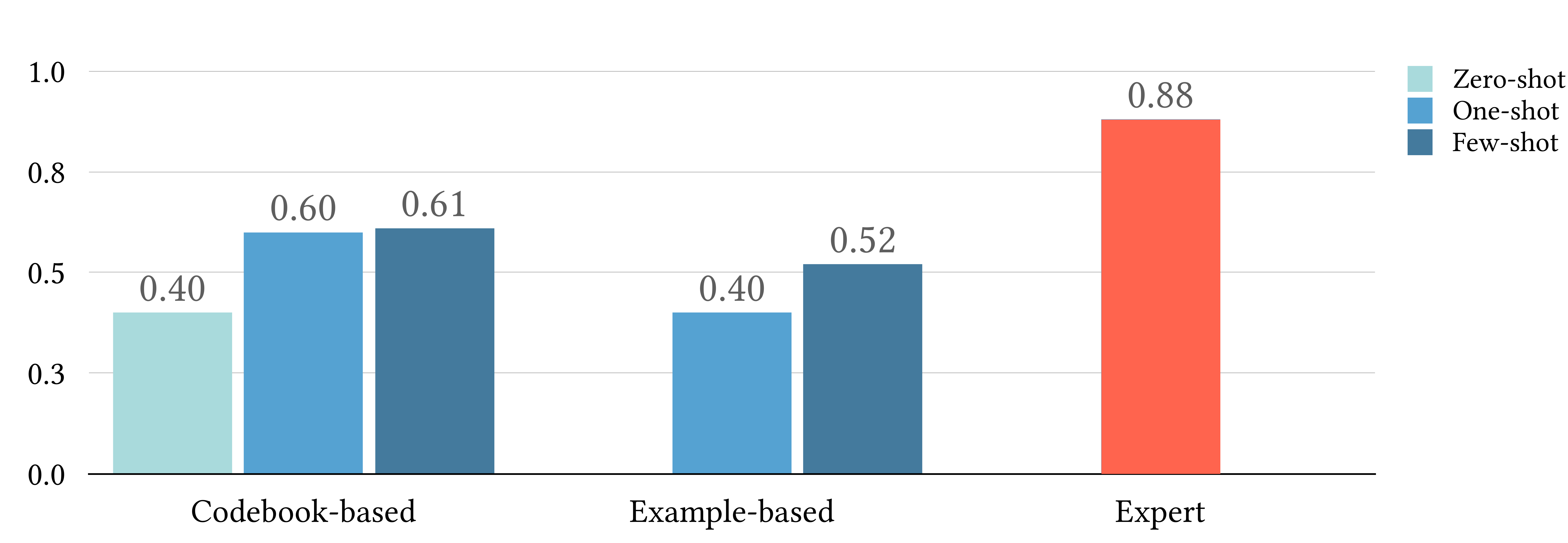}
            \caption[]%
            {{\small Cohen's $\kappa$ for Question Complexity}}    
        \end{subfigure}
         \begin{subfigure}[b]{1\textwidth}
            \centering
            \includegraphics[width=0.6\textwidth]{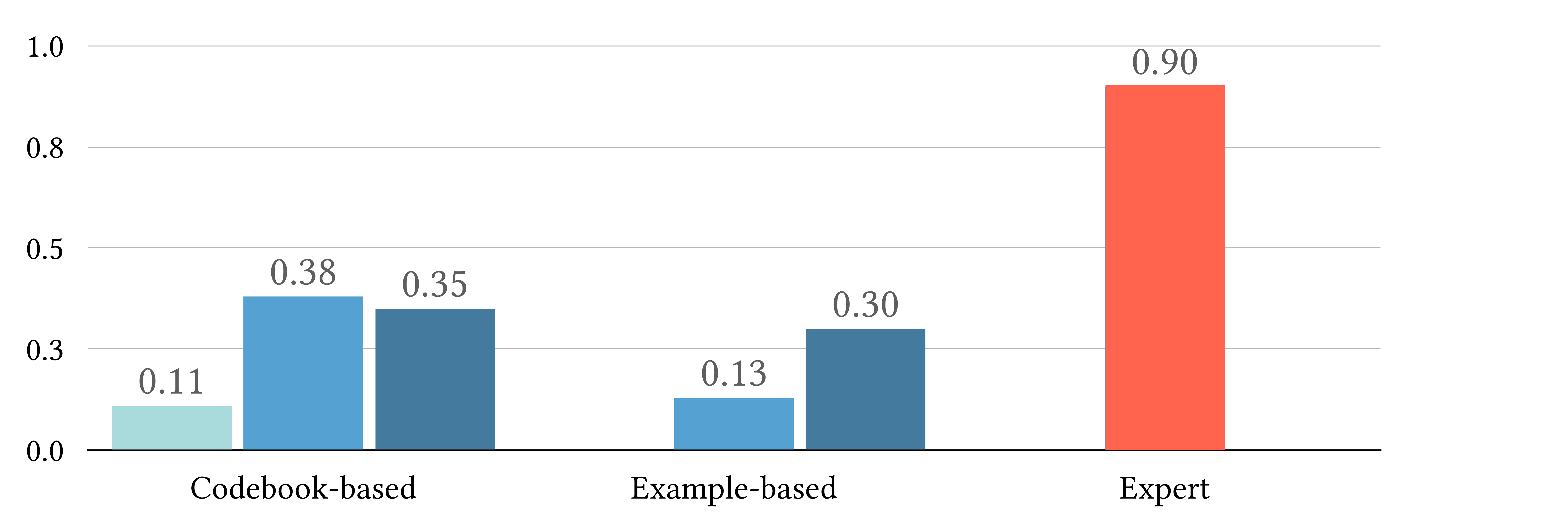}
            \caption[]%
            {{\small Cohen's $\kappa$ for Syntactic Structure}}    
        \end{subfigure}
        \caption[] %
        {\small The Cohen's $\kappa$ between GPT-3 and experts shows substantial agreement in Question Complexity coding and fair agreement in Syntactic Structure coding. In general, Codebook-centered prompts with examples achieves the highest agreement.} 
    \label{fig:figures}
    \end{figure*}
For all prompt variants, we included an identity modifier, ``I am a developmental psychologist who has expertise in linguistics.'' and an instruction to constrain output space, ``Choose from the following candidates: [Code Set]''.

\section{Results}
We measured the performance of our GPT-3 based approach with Cohen's Kappa \cite{mchugh2012interrater}. Cohen's Kappa measures inter-rater reliability, which indicates how two coders agree with each other. We computed two sets of Cohen's Kappa, between GPT-3 with the expert's final coding results and between two experts who originally coded the dataset using the \textit{same} codebook.

For RQ1, our results suggest that it is feasible to use GPT-3 with an expert-developed codebook for deductive coding. When analyzing curiosity-driven questions, our GPT-3-based approach achieved fair (Syntactic Structure: Cohen's $\kappa = 0.38$) to substantial (Question complexity: Cohen's $\kappa = 0.61$) agreement with expert rating (Cohen's $\kappa$'s interpretation is based on  \cite{mchugh2012interrater}), see Fig. \ref{fig:figures}. However, there is a gap between experts' agreement with our approach and the agreement among experts (Question complexity: Cohen's $\kappa = 0.88$; Syntactic Structure: Cohen's $\kappa = 0.90$).

For RQ2, we examined the different prompt designs. We found the codebook-centered design performs better than the example-centered designs, see Fig. \ref{fig:figures}. And examples play an important role. We observed the largest performance gain when shifting from a zero-shot to a one-shot setting. However, the performance between one-shot and few-shot settings did not differ much.

\section{Opportunities and Challenges}
Our preliminary findings indicate the feasibility of using LLMs for qualitative analysis. In a deductive coding task, by combining GPT-3 and a codebook, our LLM-based approach achieved fair to substantial agreement with experts. Considering the accessibility and flexibility of LLMs, we believe this approach has the potential to effectively help researchers to analyze qualitative data, especially for increasingly large datasets. We lay out a few challenges and opportunities for our future studies, 

\begin{itemize}
    \item Interrogate Model Capability: In this preliminary exploration, we only measured the level of agreement. To better understand model capability, we ought to conduct more detailed error analyses on disagreed items. Also, it is unclear if the LLM-based method could be extended to different contexts and different coding schemes where the coder needs to capture more nuanced signals. 

    \item Design for Appropriate Reliance: Although our preliminary results show fair to substantial agreement with expert rating, the model produces incorrect labels. When deploying such an imperfect AI system, we must design for appropriate reliance to prevent over-trusting and misuse. For example, when designing the interface, we could consider explainable AI methods to calibrate user trust over time. 
    
    \item Design Codebook for LLMs: We constructed our prompts using the codebook for experts. Although it provides transparency and explicit control, it may limit the model's performance. Future study is required to understand how to design a more effective codebook for task performance and model understanding. 
    
    \item Support Inductive Coding: We demonstrated the feasibility of using LLMs in deductive coding. However, for inductive coding, where the process is more bottom-up and exploratory, novel interaction and Human-AI collaboration diagrams are required. We should study interaction techniques and controls to let researchers use LLMs more effectively in the qualitative analysis.
\end{itemize}


\bibliographystyle{ACM-Reference-Format}
\bibliography{sample-base}


\end{document}